\documentclass[twocolumn]{svjour3}          

\usepackage{graphicx}
\usepackage{amsmath,amssymb}
\usepackage{color}
\usepackage{caption}
\usepackage{subfigure}
\usepackage{breakurl}
\usepackage{multirow}
\usepackage{booktabs}
\usepackage{setspace}
\usepackage{mathrsfs}
\usepackage{algorithm}
\usepackage{algpseudocode}
\usepackage{dsfont}
\usepackage{wasysym}
\usepackage[misc]{ifsym}
\usepackage[pagebackref=false,breaklinks=true,letterpaper=true,colorlinks,urlcolor=blue,citecolor=blue,linkcolor=blue,bookmarks=false]{hyperref}

\begin{document}

\title{Joint Face Hallucination and Deblurring via \\ Structure Generation and Detail Enhancement}


\author{Yibing Song \and Jiawei Zhang \and Lijun Gong \and Shengfeng He \and \\
Linchao Bao \and Jinshan Pan \and Qingxiong Yang \and Ming-Hsuan Yang       
}


\institute{Yibing Song and Linchao Bao \at Tencent AI Lab, Shenzhen, China.\and
Jiawei Zhang \at Sensetime Research, Shenzhen, China.\and
Lijun Gong \at Tencent, Shenzhen, China.\and
Shengfeng He \at South China University of Technology, Guangzhou, China.\and
\Letter~Jinshan Pan \at Nanjing University of Science and Technology, Nanjing, China. Corresponding author.\and
Qingxiong Yang \at University of Science and Technology of China, Hefei, China.\and
Ming-Hsuan Yang \at University of California at Merced, Merced, U.S.\\
}

\date{Received: date / Accepted: date}

\maketitle

\begin{abstract}
We address the problem of restoring a high-resolution face image from a blurry low-resolution input.
This problem is difficult as super-resolution and deblurring need to be tackled simultaneously.
Moreover, existing algorithms cannot handle face images well as low-resolution face images do not have much texture which is especially critical for deblurring.
In this paper, we propose an effective algorithm by utilizing the domain-specific knowledge of human faces to recover high-quality faces.
We first propose a facial component guided deep Convolutional Neural Network (CNN) to restore a coarse face image, which is denoted as the base image where the facial component is automatically generated from the input face image.
However, the CNN based method cannot handle image details well.
We further develop a novel exemplar-based detail enhancement algorithm via facial component matching.
Extensive experiments show that the proposed method outperforms the state-of-the-art algorithms both quantitatively and qualitatively.
\keywords{Face Hallucination \and Face Deblurring \and Convolutional Neural Network}
\end{abstract}

\section{Introduction}
\begin{figure*}[t]
\begin{center}
\begin{tabular}{ccccc}
\includegraphics[width=0.18\linewidth]{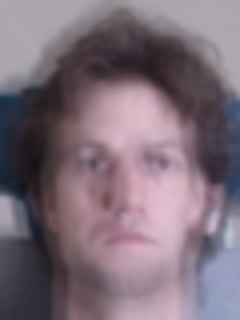}&
\includegraphics[width=0.18\linewidth]{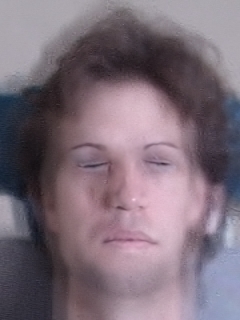}&
\includegraphics[width=0.18\linewidth]{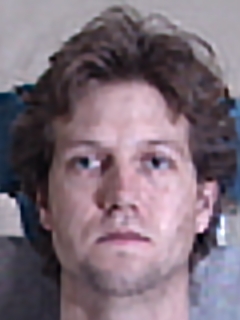}&
\includegraphics[width=0.18\linewidth]{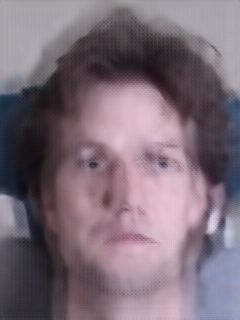}&
\includegraphics[width=0.18\linewidth]{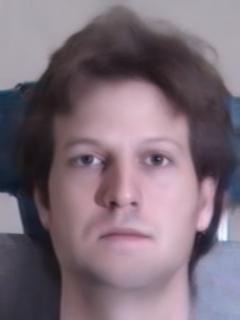}\\
(a) Input (Bic)&(b) SFH \cite{yang-cvpr13-sfh}&(c) DFE \cite{pan-eccv14-dfe}&(d) RBF \cite{xu-iccv17-learning}&(e) Ours
\end{tabular}
\end{center}
\caption{Joint face hallucination and deblurring. The input blurry LR image (with bicubic upsampling) is shown in (a). The results generated by existing face hallucination and deblurring methods are shown from (b) to (d). Different from existing methods which treat face hallucination and deblurring separately, we formulate these two tasks into a single framework for joint prediction. Our method effectively recovers the facial components as shown in (e) and performs favorably against the state-of-the-art methods.}
\label{fig:intro}
\end{figure*}

Human faces captured in the real world usually suffer from the imaging process.
The large distance between human faces and camera leads to limited pixel sampling on the camera sensor.
Meanwhile, relative movement during exposure brings blur to the digital images.
As a result, the captured faces are usually in small resolution and contain moderate blur.
As camera viewpoints cannot be changed frequently in some cases (e.g., video surveillance scenario), there is a need to restore the face images for further analysis.
However, existing algorithms designed for image super-resolution or image deblurring cannot handle this problem well due to the influence of both resolution and blur (Figure~\ref{fig:intro}).
As human faces contain rich details around facial components, i.e., eyes, mouth, etc., it is of great interest to develop an effective algorithm to estimate clear high-resolution (HR) face images by the domain-specific knowledge of faces.

\renewcommand{\tabcolsep}{1pt}

When handling the restoration problem of face images, the state-of-the-art face hallucination (FH) methods usually learn the mapping function from low-resolution (LR) images to high-resolution HR images
either in a regression~\cite{liu-ijcv07-face,jia-tip08-generalized,tappen-eccv12-bayesian} or an exemplar-based~\cite{ma-pr10-hallucinating,jia-cvpr06-multi,jia-iccv05-multi} way.
Although existing FH algorithms achieve great progress, these algorithms usually assume the blur is intrinsic (e.g., bicubic blur, Gaussian blur). Meanwhile, they are less effective when the inputs contain heavy motion blur.
Several algorithms~\cite{pan-eccv14-dfe} have been proposed to deal with blurry face images. However, these algorithms usually assume the high resolution of the blurry input. If the resolution of the blurry input faces is low, these algorithms are not able to generate reliable data for blur kernel estimation~\cite{pan-eccv14-dfe}.
To solve LR blurry images, recent methods~\cite{Park_2017_ICCV} jointly estimate image super-resolution and image deblurring. Xu et al. \cite{xu-iccv17-learning} use Generative Adversarial Nets (GANs) to super-resolve face images.
However, without exploiting the unique structures of human faces, these methods are not able to handle face image restoration problem well, especially around facial components.
Figure \ref{fig:intro} shows an example where the input is a low-resolution face image.
Both the state-of-the-art FH algorithm \cite{yang-cvpr13-sfh} and face deblurring (FD) algorithm~\cite{pan-eccv14-dfe} cannot effectively recover clear HR images.
Meanwhile, the deep learning based method~\cite{xu-iccv17-learning} does not effectively recover the detailed structure and reduce blur.

In this paper, we propose a unified framework to super-resolve face images.
As this is an ill-posed problem and most information is missing,
recovering a face image with both general facial structure and local details using one CNN framework without any domain knowledge is challenging.
We formulate the restored image using as a base layer and a detail layer.
The base layer is learned by using a CNN guided by facial components. The detail layer is generated by
an exemplar-based texture synthesis module.
First, our facial structure generation network (FSGN) takes the up-sampled face image and its facial components as the inputs and generates base images. Then we use a patch-wise K-Nearest Neighbor (K-NN) to search between the intermediate face image and exemplar images. In this way, we can accurately establish the correspondences on the HR training images and overcome the limitations of existing feature matching-based methods. The accurate correspondence ensures that the fine-grained facial structures from the HR exemplar images are effectively extracted.
Finally, the details from these structures are transferred into the base image through edge-aware image filtering.
Figure \ref{fig:intro}(e) shows that our algorithm is able to super-resolve blurry face images and generates the face image with much more clear textures.

The contributions of this work are summarized as follows:
\begin{itemize}
  \item We propose a unified framework for joint face hallucination and deblurring by using the special properties of face images. To generate high-quality faces, we develop a face component guided CNN.
  \item We develop a novel exemplar-based detail transfer algorithm to improve the details and texture estimations of face images.
  \item We analyze the properties of the proposed algorithm and show that it performs favorably against state-of-the-art face hallucination and deblurring methods on the public benchmarks.
\end{itemize}

\section{Related Work}
Face hallucination and deblurring relate closely to generic image super-resolution and deblurring. In this section, we perform a literature review on the most related work of face hallucination, image super-resolution, and face deblurring and put this work in proper context.

\subsection{Face Hallucination}
Learning based methods are widely adopted in face analysis approaches including face hallucination \cite{wang-ijcv14-survey,song-eccv14-sketch,song-ijcai17-faceSketch} and style transfer \cite{song-cviu17-stylizing}.
Face hallucination methods can be categorized as the data-driven framework and the CNN generative framework.
In the data-driven framework, various approaches are proposed to learn the transformation between LR and HR to recover the missing details from the input.
In \cite{gunturk-tip03-eigenface,wang-SMC05-hf}, generalized approaches on the eigen domain are proposed to map both LR and HR image spaces.
The tensor-based methods are proposed by \cite{liu-cvpr05-hf,jia-tip08-generalized} to well hallucinate multiple model face images across different poses and expressions.
In \cite{liu-ijcv07-FH}, Principle Component Analysis (PCA) based linear constraints are learned from the training images and a patch-based Markov Random Field (MRF) is used to reconstruct the residues. It can only work on fixed poses and expressions.
In \cite{Jin-cvpr15-FH}, the blurring kernel and transformation of LR faces are jointly estimated by deblurring and registration in PCA subspace.
It only works for face region instead of the whole face image. Face hallucination in the compressed domain is proposed in \cite{liu-icip14-fh,yang-ijcv17-fh}.
Image alignment-based methods are adopted for face hallucination where HR face images are matched to LR face images by dense SIFT flow \cite{tappen-eccv12-bayesian} or feature matching \cite{yang-cvpr13-sfh,song-ijcai17-faceSR}.
The quality of output HR results depends on image alignment which is less effective when poses and expressions are different between training and input images.

On the other hand, the CNN generative framework predicts HR face images in an end-to-end manner.
In \cite{zhou-aaai15-learning}, a Bi-channel CNN is proposed to integrate the input image and face representation for prediction. In \cite{yu-eccv16-ultra,Tero-iclr18-gan}, the GAN framework is applied to hallucinate LR face images.
However, the network generates high-resolution images from random noise in \cite{Tero-iclr18-gan} while the face hallucination task is to tackle a specific input image.
The transformative discriminative auto-encoders are proposed in \cite{yu-cvpr17-hallucinating,chen-arxiv17-fsrnet,jourabloo-iccv17-pose} to upsample images and denoise simultaneously during the hallucination.
In \cite{yu-aaai17-face}, a spatial alignment network is proposed for LR and HR matching.
A cascaded bi-network is proposed in \cite{zhu-eccv16-deep} for FH and deep reinforcement learning is applied in \cite{cao-cvpr17-attention} to achieve attention awareness.
The CNN generative framework usually handles input face images in an extremely low resolution where the facial components are not able to be distinguished.
Even though these methods generate facial structures on the output HR result, these structures are not accurate and lead to incorrect identity.
In contrast, our method combines the advantage of both CNN generative and data-driven framework for identity preservations.

\subsection{Image Super Resolution}
The advancement of CNN has activated a series of investigations on image SR.
Starting from SRCNN \cite{dong-eccv14-srcnn,dong-pami2016-image} where HR images are predicted in an end-to-end manner
via several convolutional layers and nonlinear activations, following works are proposed to combine existing models with CNN \cite{yang-cvpr18-image} or improve network capacity.
In \cite{wang-iccv15-deep}, a sparse coding model is designed for SR and incarnated as a neural network, which is trained end-to-end in a cascade structure.
A convolutional sparse coding method is proposed in \cite{shuhang-iccv15-csc} to enforce the pixel consistency during image reconstruction.
It exploits the image global correlation to produce a more robust reconstruction of image local structures.
In \cite{shi-cvpr16-real}, an efficient sub-pixel convolutional layer is proposed to learn an array of upscaling filters for SR.
It will replace the handcrafted upscaling filters for each CNN feature map specifically.
A deeply-recursive convolutional network is proposed in \cite{kim-cvpr16-deeply} to involve recursion, whose depth can improve performance without introducing new parameters for additional convolutions.
In \cite{kim-cvpr16-accurate}, a very deep convolutional network is designed to increase network depth and residual learning is involved to facilitate the training process. An accelerated version of SRCNN is proposed in \cite{dong-eccv16-accelerating} to achieve real-time performance for practical usage. In \cite{johnson-eccv16-perceptual}, a perceptual loss is introduced for training feed-forward networks in image SR.
The generative adversarial network (GAN) is applied to image SR in \cite{ledig-cvpr17-photo} to achieve perceptually satisfaction.
In \cite{lai-cvpr17-deep}, a deep laplacian pyramid network is proposed to upsample LR images gradually via deconvolution. Meanwhile, residual learning \cite{tai-cvpr17-image} is involved to facilitate the training process.
Anchored regression network is proposed in \cite{agustsson-iccv17-anchored} to design a smoothed relaxation of a piecewise linear regressor through the combination of multiple linear regressors over soft assignments to anchor points. In \cite{sajjadi-iccv17-enhancenet}, texture synthesis is proposed in combination with perceptual loss focusing on creating realistic textures rather than optimizing pixel-wise loss function.
In~\cite{timofte-cvprw17-ntire}, it has been shown that the method by Lim et al. \cite{lim-cvprw17-edsr} performs well against the sate-of-the-art SR algorithms.
Different from existing image SR methods, our algorithm is specifically designed for FH and focus on facial structure generation and enhancement.

\begin{figure*}[t]
\begin{center}
\begin{tabular}{c}
\includegraphics[width=0.95\linewidth]{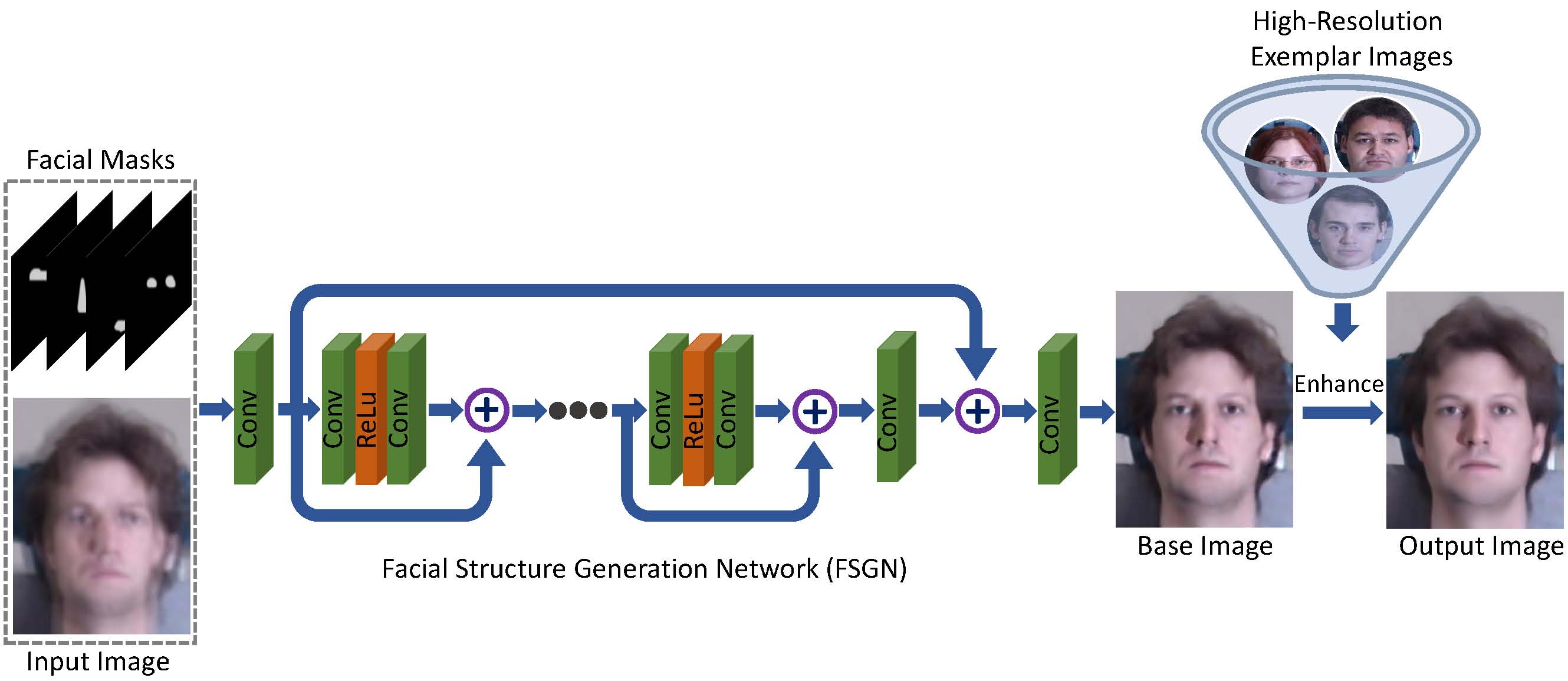}\\
\end{tabular}
\end{center}
\caption{Pipeline of the proposed method. Given a blurry LR input image, we first upsample it via bicubic interpolation and obtain the facial landmarks for generating facial components. Then we develop a facial structure generation network  to generate the base image. Finally, we develop a details enhancement algorithm to estimate the missing details in the base image by HR exemplar images.}
\label{fig:pipeline}
\end{figure*}

\subsection{Face Deblurring}
Most existing deblurring algorithms~\cite{pan-pami17-deblurring,pan-2017-learning,pan-pami17-deblurring,zhang-cvpr18-dynamic} focus on the generic image deblurring. As blurry face images contain fewer textures, the generic deblurring algorithms cannot handle this problem well.
In \cite{hacohen-iccv13-deblurring}, a non-rigid dense correspondence is established and blur kernel estimation is performed accordingly.
Facial structures are exploited in \cite{pan-eccv14-dfe} to maximum a posteriori deblurring algorithm on an exemplar dataset.
These data-driven methods are able to handle blurry images when sharp edges are extracted effectively while requires querying time cost.
The deep learning algorithm has also been applied to face deblurring. In~\cite{xu-iccv17-learning}, Xu et al. use GAN to super-resolve face images. It predicts face images via extremely low-resolution inputs where the facial identity is hard to identify.
However, as this method does not consider the specialty of face images, it is less effective to restore some key component of faces.
%
Recently, Shen et al.~\cite{shen-cvpr18-deep} propose a face deblurring method which uses
several CNNs to generate semantic face labels for guiding the face deblurring process.
In this work, we solve the hallucination and deblurring in a unified framework.
We note that  Shen et al.~\cite{shen-cvpr18-deep} use computationally expensive face parsing CNNs to generate pixel-wise semantic face labels.
In addition, the prediction accuracy decreases when the input resolution is low, and thus affects the following process.
In contrast, we use a computationally efficient facial landmark detector to estimate the facial components to
guide the FSGN.
Our experimental results show that the proposed facial landmark performs robustly against when the landmarks are not precisely detected.
Shen et al.~\cite{shen-cvpr18-deep} use two cascaded CNNs to reduce the blurry effect on the face images while we use the FSGN to generate a base image containing the general facial structure and
an exemplar-based texture synthesis framework to restore details.

\begin{table}[t]
\renewcommand\arraystretch{1.2}
\begin{center}
\begin{tabular}{|c|c|}
  \hline
  layer name & FSGN \\ \hline
  conv 1 & $3\times 3$, 64, pad 1 \\ \hline
  conv 2x & $\left[\begin{array}{c}3\times3,~64,~\rm{pad}~5,~\rm{dilation}~5\\3\times3,~64,~\rm{pad}~5\end{array} \right]\times5$\\ \hline
  conv 3x & $\left[\begin{array}{c}3\times3,~64,~\rm{pad}~4,~\rm{dilation}~4\\3\times3,~64,~\rm{pad}~4\end{array} \right]\times5$\\ \hline
  conv 4x & $\left[\begin{array}{c}3\times3,~64,~\rm{pad}~3,~\rm{dilation}~3\\3\times3,~64,~\rm{pad}~3\end{array} \right]\times5$\\ \hline
  conv 5x & $\left[\begin{array}{c}3\times3,~64,~\rm{pad}~2,~\rm{dilation}~2\\3\times3,~64,~\rm{pad}~2\end{array} \right]\times5$\\ \hline
  conv 6x & $\left[\begin{array}{c}3\times3,~64,~\rm{pad}~1,~\rm{dilation}~1\\3\times3,~64,~\rm{pad}~1\end{array} \right]\times5$\\ \hline
  conv 7 & $3\times3$, 64, pad 1\\ \hline
  conv 8 & $3\times3$, 64, pad 1\\ \hline
\end{tabular}
\end{center}
\caption{Network parameters of the proposed facial structure generation network.}
\label{tab:arch}
\end{table}
\vspace{-3mm}

\section{Proposed Method}
Figure \ref{fig:pipeline} shows the pipeline of our method. It mainly consists of two modules.
The first module is a very deep CNN, namely Facial Structure Generation Network, which predicts a base image given the LR input.
The base image contains the basic structure of the input face while the facial details are not fully recovered.
It is then fed into the second module for detail enhancement.
Note that the second module of our method relies on establishing correspondences from the base image to high-resolution exemplar images, which benefits from the first module since major structures of the input face are roughly recovered by the FSGN and thus the establishment of LR-HR correspondences are much easier.
We describe the details of the two modules in the following:

\subsection{Facial Structure Generation Network}
As the unique structure of human faces differs much from the natural images (i.e., textures mostly reside around the facial components), we expect our FSGN to focus on the facial components rather than the remaining regions which are typically flat and less informative. Given an input LR blurry face image, we first upsample it using bicubic interpolation to the same resolution of the output.
Then facial landmarks are detected on the upsampled image using the method from \cite{zhu-cvpr12-face}. Facial component masks are then generated using the landmarks.
As the input LR images contain blurry pixels, the facial masks may not accurately localize the facial components.
Nevertheless, our FSGN performs well when the facial masks cannot be accurately extracted.
Following \cite{yang-cvpr13-sfh}, we categorize facial components into four types, which are eyebrows, eyes, noses, and mouths, respectively. For each type of facial component, we prepare a mask where the pixels within the component region are marked as 1 and the others as 0. In total, we generate four masks covering all four types of facial components accordingly. Figure \ref{fig:pipeline} shows a direct view of these masks.

\subsubsection{Network Architecture}
The FSGN consists of 25 residual blocks \cite{he-cvpr16-deep} with 53 convolution layers in total.
In addition to the local skip connections in each residue block, we add an additional long-range skip connection from the first convolution layer to the last convolution layer.
The network is fully convolutional and we use a dilated $3 \times 3$ convolution \cite{yu-cvpr17-dilated} in all the layers.
There is no pooling layer in our network and the size of all intermediate feature maps is the same as the input image.
The detailed network architecture is shown in Table \ref{tab:arch}.
During training, we create low-resolution blurry images from the ground-truth high-resolution images for input. Meanwhile, these corresponding ground-truth images are used for supervision with the Euclidean loss.

\subsection{Detail Enhancement}
\begin{figure*}[t]
\begin{center}
\begin{tabular}{cccc}
\includegraphics[width=0.24\linewidth]{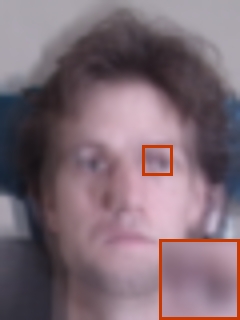}&
\includegraphics[width=0.24\linewidth]{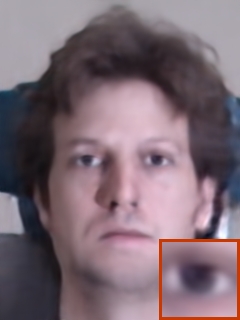}&
\includegraphics[width=0.24\linewidth]{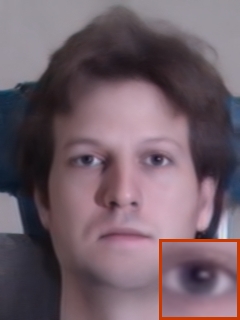}&
\includegraphics[width=0.24\linewidth]{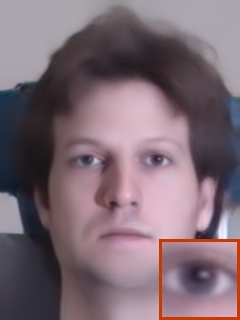}\\
\small{(a) Input}&\small{(b) Base image}&\small{(c) Exemplar regression}&\small{(d) GF on (b) guided by (c)}\\
\includegraphics[width=0.24\linewidth]{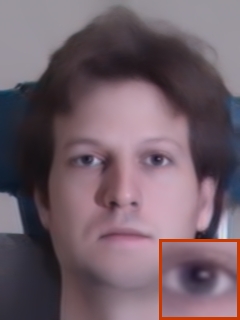}&
\includegraphics[width=0.24\linewidth]{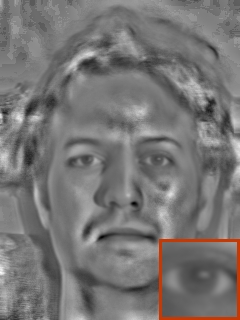}&
\includegraphics[width=0.24\linewidth]{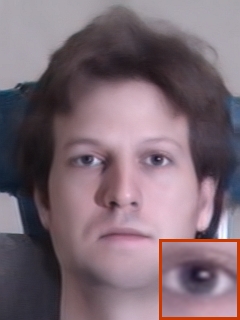}&
\includegraphics[width=0.24\linewidth]{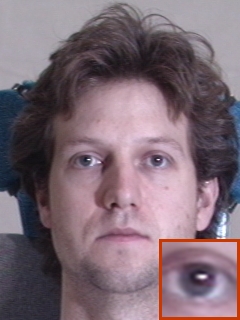}\\
\small{(e) GF on (c) guided by (c)}&\small{(f) Detail: (c)-(e)}&\small{(g) Output: (d)+(f)}&\small{(h) Ground Truth}\\
\end{tabular}
\end{center}
\caption{The process of structure enhancement. The input LR blurry face image is shown in (a). The base image generated by FSGN is shown in (b). The exemplar regression result is shown in (c). We perform guided filtering on (b) using (c) as guidance to get (d). We also filter (c) using guided filtering in (e). The lost structure details after filtering are shown in (f), which is the difference between (c) and (e). We add the details back to (d) to generate the output as shown in (g). The ground truth image is shown in (h).}
\label{fig:structure_transfer}
\end{figure*}

Although the proposed FSGN is able to restore an image in which major facial components can be effectively recovered, it tends to over-smooth the details of recovered face images (as shown in Figure~\ref{fig:structure_transfer}(b)).
To solve this problem, we propose a detail enhancement method to estimate the missing details by using high-resolution exemplar images.
Our detail enhancement method consists of two steps.
In the first step, we establish the patch correspondences between the base image generated by FSGN and HR exemplar images, then we use the patches from HR exemplar images to regress the base image and get an intermediate result.
In the second step, the details from the intermediate result are transferred to the base image via edge-preserving filtering to obtain the final result.
Figure \ref{fig:structure_transfer} illustrates the effectiveness of our detail enhancement method. The details are presented in the following sections.

\subsubsection{Exemplar Regression}
Given the base image produced by FSGN, we divide it into local patches. For one patch centered on pixel $p$, we perform a $K$ nearest neighbor search ($K$-NN) in the HR exemplar images to find the $K$ most similar patches. Note that the HR exemplar images are face images where the subjects are different from that in the input image.
In the $K$-NN patch search step, we choose a search region in one HR exemplar image for each input patch.
The center of each search region is the same as that of the input patch.
In the search region, we use a sliding window to select one patch, which is the same size as the input patch.
We obtain $N$ patches from $N$ training HR images after patch search and further select $K$ among them.
\begin{equation}
D_p=\alpha\cdot(1-D_{ncc})+(1-\alpha)\cdot D_{abs},
\label{eq:ncc_abs}
\end{equation}
where $\alpha$ is the weight combining the two metrics. It is set as 0.5 in our implementation.
We normalize image pixel value to $[0,1]$ in order to set the two metrics into the same range.

After $K$-NN search we select $K$ candidate patches from HR exemplar images. Let $\mathrm{H}_p^i$ ($i\in[1,\cdots,K]$) denote a vector containing all the pixel values of the $i$th HR candidate patch, and $\mathrm{I}_p$ denote a vector containing the pixel values of the input patch. We also denote the linear regression function as $\mathcal{F}_p=[F_p^1,\cdots,F_p^K]^\mathrm{T}$ where $F_p^i$ ($i\in[1,\cdots,K]$) is each coefficient of $\mathcal{F}_p$. The energy function is defined as:
\begin{equation}
E_p^{\mathrm{data}}=||\mathrm{H}_p\cdot\mathcal{F}_p-\mathrm{I}_p||^2,
\label{eq:energy_data}
\end{equation}
where $\mathrm{H}_p=[\mathrm{H}_p^1,\mathrm{H}_p^2,\cdots,\mathrm{H}_p^K]$. It is a linear regression form and we can compute $\mathcal{F}_p$ as
\begin{equation}
\mathcal{F}_p=(\mathrm{H}_p^\mathrm{T}\cdot\mathrm{H}_p)^{-1}\mathrm{H}_p^\mathrm{T}\cdot\mathrm{I}_p.
\label{eq:solver}
\end{equation}
We can efficiently compute $\mathcal{F}_p$ when the patches contain texture (i.e., the pixel values in $\mathrm{H}_p$ should not be similar to each other). However, in some cases when $p$ is in the smooth region (e.g., cheek) $\mathrm{H}_p^\mathrm{T}\cdot\mathrm{H}_p$ may become a singular matrix and thus $\mathcal{F}_p$ is not accurate. We resolve the problem by adding a regularization term as:
\begin{equation}
E_p = E_p^{\mathrm{data}}+E_p^{reg} = ||\mathrm{H}_p\cdot\mathcal{F}_p-\mathrm{I}_p||^2+\lambda ||\mathcal{F}_p||^2,
\end{equation}
where $\lambda$ is the weight controlling the influence of regularization term. It is set as the number of pixels in an input patch. We can solve the above energy function as:
\begin{equation}
\mathcal{F}_p=(\mathrm{H}_p^\mathrm{T}\cdot\mathrm{H}_p+\lambda\mathds{1})^{-1}\mathrm{H}_p^\mathrm{T}\cdot\mathrm{I}_p,
\label{eq:solver_final}
\end{equation}
where $\mathds{1}$ is the identity matrix.

Once we calculate the regression function $\mathcal{F}_p$, we map the HR exemplar patches into the output patch. Let $\bar{\textrm{H}}_p^i$ ($i\in[1,\cdots,K]$) denote one vector containing the pixel values of the corresponding HR exemplar patches. The output patch $\textrm{R}_p$ can be computed as:
\begin{equation}
\textrm{R}_p=\sum_{i=1}^K{\bar{\textrm{H}}_p^i\cdot F_p^i}.
\label{eq:mapping}
\end{equation}
We compute the output patch for each pixel. For the overlapping areas between different patches, we perform averaging to generate the final regression result.

\begin{algorithm}[t]
\caption{Overview of proposed algorithm}
\begin{algorithmic}[1]
\State \emph{-- Training --}
\State Train FSGN using face images and masks;
\State \emph{-- Testing (input LR blurry image $\mathbb{I}$) --}
\State Generate base image $\bar{\mathbb{I}}$ using FSGN;
\For{each pixel $p$ in $\bar{\mathbb{I}}$}
    \State $K$-NN patch search using equation \ref{eq:ncc_abs};
    \State Calculate regression matrix using equation \ref{eq:solver_final};
    \State Perform regression $R_p$ using equation \ref{eq:mapping};
\EndFor
\State Guided filtering on $\bar{\mathbb{I}}$ using regression image $\mathbb{R}$ to obtain $\bar{\mathbb{I}}^\mathbb{R}$;
\State Guided filtering on $\mathbb{R}$ using regression image $\mathbb{R}$ to obtain $\mathbb{R}^\mathbb{R}$;
\State Compute the output image by $\bar{\mathbb{I}}^\mathbb{R}+\mathbb{R}-\mathbb{R}^\mathbb{R}$.
\end{algorithmic}
\label{algo:code}
\end{algorithm}

\subsubsection{Detail Transfer}
The regression result contains detailed structures transferred from HR exemplar images.
However, it cannot be directly adopted as the output. This is because the detailed structures are transferred from exemplar patches which belong to different subjects.
The lighting condition of each subject is different from each other, which results in different shading appearances in facial regions between the regressed image and the ground truth (e.g., Figure \ref{fig:structure_transfer}(c) and (h)).
We here present an algorithm based on joint edge-preserving filtering \cite{petschnigg-sig04-digital,eisemann-sig04-flash} to combine the low-frequency appearances of the base image and the high-frequency facial details of the regressed image to generate the final output.

The main steps of our algorithm are shown in Figure \ref{fig:structure_transfer}.
We have a base image shown in (b) and the regressed image shown in (c).
We use guided filter \cite{he-eccv10-guided,he-pami13-guided} to smooth (b) using (c) as guidance.
As such, the facial details of (c) can be transferred into (b). However, the filtered result is likely to be over-smoothed (as shown in Figure \ref{fig:structure_transfer}(d)).
Nevertheless, we can further extract details from (c) and add them to the filtered result.
Specifically, we smooth (c) using guided filtering with itself as guidance shown in (e).
Then the smoothed details can be captured by subtracting the smoothed image (e) from (c), as shown in (f).
Finally, we add (f) to (d) to get the output image shown in (g).
Note that both global appearances and facial details of the output image are similar to the ground truth shown in (h).
The pseudo code of our entire algorithm is shown in Algorithm \ref{algo:code}.

Our detail enhancement method improves the base image quality by adding identity-specific details.
Figure \ref{fig:structure_transfer} shows that in (b) only the general facial structure is recovered in the base image while the details are still missing.
We use the exemplar regression to synthesize the details specifically for the input image.
The details are then extracted and transferred to the base image using the guided filter shown in (g).
Compared with the base image, our detail enhancement method enriches the local details around facial components while the artifacts are not involved.

\section{Experimental Results}\label{sec:experiments}

We conduct experiments on the Multi-PIE \cite{gross-ivc10-multiPie} and PubFig \cite{kumar-iccv09-pubFig} datasets. The face images in the Multi-PIE dataset are taken in the lab controlled environment while the face images in the PubFig dataset are taken in the real world condition. The resolution of the ground truth images in these two datasets is 320$\times$240. We evaluate our method from two aspects. First, we conduct an ablation study to illustrate the effectiveness of our modules. Second, we compare our method with the state-of-the-art FH methods including FHTP~\cite{liu-ijcv07-FH}, SFH~\cite{yang-cvpr13-sfh}, five image SR methods including bicubic interpolation, SRCSC~\cite{shuhang-iccv15-csc}, SRCNN~\cite{dong-pami2016-image}, VDSR~\cite{kim-cvpr16-accurate}, SRResNet~\cite{ledig-cvpr17-photo}, and two face deblurring methods DFE \cite{pan-eccv14-dfe}, RBF \cite{xu-iccv17-learning}. We use PSNR and SSIM \cite{wang-tip04-SSIM} to quantitatively measure the image quality of the generated results.

{\flushleft \bf{Training data and configurations.}} We follow the same setting with that in SFH~\cite{yang-cvpr13-sfh} and use 2184 images from the Multi-PIE dataset as training data. To create the input LR blurry images, we first convolve with the ground truth images using random blur kernel and downsample the convolved results. The blur kernel size ranges randomly from 11 to 31, and the Gaussian variance ranges randomly from 1.4 to 1.7. In total, we have generated 200 motion blur kernels without noise and randomly select one to convolve with the ground truth images. The scaling factor is set to 4. To create the input LR images without blur, we convolve with the ground truth images using Gaussian blur kernel and downsample to generate the training inputs. We train our network from scratch using the ADAM solver \cite{kingma-iclr14-adam} with a learning rate of 1e-4.
For performance evaluation against the state-of-the-art methods,
we follow the original network architecture designs and train them from scratch. The training data is the same as ours and we follow their training configurations to reproduce the results for comparison.

{\flushleft \bf{Test data.}} There are 342 test images from the Multi-PIE dataset and 400 images from the PubFig dataset, respectively. We create input images through convolving with test images via random blur kernel and Gaussian kernel. The test images are generated in the same way as training image inputs. Note that there is no identity overlap between the training and test images.

\subsection{Ablation Studies}

\def\pp{\hspace{0mm}}
\renewcommand{\tabcolsep}{6pt}
\begin{table}[t]
\caption{Experimental results using different patch sizes during $K$-NN search on the Multi-PIE dataset.}
\centering
       \begin{tabular}{ccc}
        \toprule
        &PSNR&SSIM\\
        Patch Size&SR / Deblur &SR / Deblur\\
        \midrule
        $10\times10$&34.14 / 24.62&0.91 / 0.83\\
        $15\times15$&34.52 / 24.87&0.91 / 0.84\\
        $20\times20$&\textbf{34.93} / \textbf{25.75}&\textbf{0.92} / \textbf{0.86}\\
        $25\times25$&34.65 / 25.43&0.92 / 0.85\\
        $30\times30$&34.42 / 25.21&0.91 / 0.85\\
        \bottomrule
       \end{tabular}
\label{tab:ablation_patchsize}
\vspace{5mm}
\caption{Experimental results using different patch numbers during $K$-NN search on the Multi-PIE dataset.}
\centering
       \begin{tabular}{ccc}
        \toprule
        &PSNR&SSIM\\
        Patch Number&SR / Deblur &SR / Deblur\\
        \midrule
        3&34.05 / 24.97&0.87 / 0.83\\
        4&34.68 / 25.34&0.90 / 0.85\\
        5&\textbf{34.93} / \textbf{25.75}&\textbf{0.92} / \textbf{0.86}\\
        6&34.91 / 25.74&0.92 / 0.86\\
        7&34.92 / 25.75&0.92 / 0.86\\
        \bottomrule
       \end{tabular}
\label{tab:ablation_patchnumber}
\end{table}

In our detail enhancement step, we use $K$-NN search on the HR training images to establish the correspondence for exemplar regression.
As there exist several parameters, e.g., patch size and the number of candidate patches, in $K$-NN search, we analyze the effect of these parameters and show how they affect the proposed algorithm.
Table \ref{tab:ablation_patchsize} and Table \ref{tab:ablation_patchnumber} show the evaluation results. In the detail enhancement step, we set different patch sizes ranging from $10\times10$ to $30\times30$ incremented by 5.
Table \ref{tab:ablation_patchsize} shows that the proposed method performs well when the patch size is $20\times20$.
Table \ref{tab:ablation_patchnumber} shows the effect of the proposed method with different candidate patch numbers.
The proposed method performs well when the patch number is $5$.

\def\pp{\hspace{0mm}}
\renewcommand{\tabcolsep}{3pt}
\begin{table}[t]
\caption{Experimental results with average facial mask deviation on the Multi-PIE dataset. The deviation extent is measured with pixels.}
\centering
       \begin{tabular}{ccccccc}
        \toprule
        Mask Deviation&0&2&4&6&8&10\\
        \midrule
        PSNR&25.54&25.52&25.33&25.10&24.53&24.12\\
        SSIM&0.85&0.85&0.84&0.83&0.81&0.78\\
        \bottomrule
       \end{tabular}
\label{tab:ablation_mask}
\end{table}

\renewcommand{\tabcolsep}{1pt}
\begin{figure*}[t]
\begin{center}
\begin{tabular}{ccccc}
\includegraphics[width=0.19\linewidth]{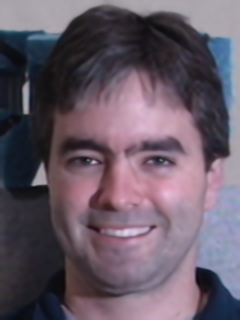}&
\includegraphics[width=0.19\linewidth]{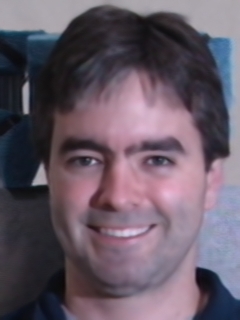}&
\includegraphics[width=0.19\linewidth]{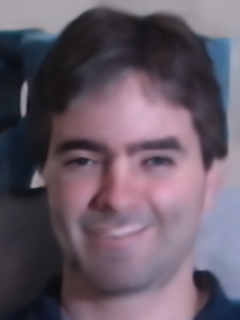}&
\includegraphics[width=0.19\linewidth]{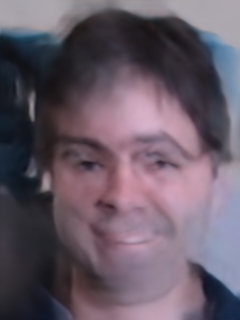}&
\includegraphics[width=0.19\linewidth]{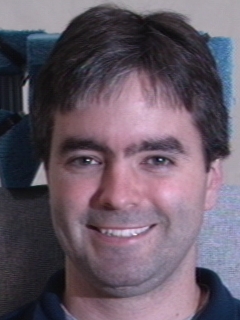}\\
\small{(a) Mask deviation 0}&\small{(b) Mask deviation 6}&\small{(c) Mask deviation 8}&\small{(d) Mask deviation 10}&\small{(e) Ground Truth}\\
\end{tabular}
\end{center}
\vspace{-3mm}
\caption{Restoration results using different facial mask deviations. The restoration result generated using ground truth facial mask is shown in (a). The restoration results generated by less accurate facial masks are shown from (b)-(d). The ground truth image is shown in (e).}
\label{fig:mask_variance}
\end{figure*}

We note that the proposed algorithm still performs well when the facial masks do not align the ground truths. In the deblurring task, we note that the facial masks may deviate according to the LR blurry input images. The deviation of the facial masks correlates with the blur kernel size. To analyze how the deviation of the facial masks affect the output result, we use different blur kernels to generate different sets of input images shown in Table \ref{tab:ablation_mask}. For each set , we generate the corresponding facial masks and compute their average deviations with the ground truth masks. The deviation is measured in pixels. Then we generate the output results for each input image and quantitatively evaluate their performance in Table \ref{tab:ablation_mask}. It shows that the performance gradually decreases as there are more deviations of the facial masks. Meanwhile, we observe that there is an obvious degradation of the output quality when the deviation exceeds 6 pixels. Figure \ref{fig:mask_variance} shows a visual example of the mask deviation. We use different blur kernel size to produce several input images and generate our results accordingly. The results indicate that when the blur kernel exceeds 31 (i.e., the mask deviation is above 6 pixels) the artifacts occur on the output images. To ensure the facial masks effective for the input images, we set the blur kernel size below 31 pixels to generate the output result.

\def\pp{\hspace{0mm}}
\renewcommand{\tabcolsep}{6pt}
\begin{table}[t]
\caption{Ablation studies on the Multi-PIE dataset with predefined five configurations. We denote baseline as B, baseline with local residual blocks as BL, facial masks input as BL+M, dilation integration as BLD+M, and detail enhancement integration as BLD+M+DE.}
\centering
       \begin{tabular}{ccc}
        \toprule
        &PSNR&SSIM\\
        &SR / Deblur &SR / Deblur\\
        \midrule
        B&33.90 / 23.97&0.87 / 0.81\\
        BL&33.95 / 24.02&0.88 / 0.82\\
        BL+M&34.22 / 24.83&0.89 / 0.84\\
        BLD+M&34.63 / 25.31&0.91 / 0.85\\
        \textbf{BLD+M+DE}&\textbf{34.93 / 25.75}&\textbf{0.92 / 0.86}\\
        \bottomrule
       \end{tabular}
\label{tab:ablation}
\end{table}

Our method consists of several modules. Our input is the LR blurry image with four facial masks. Our network structure is FSGN with dilation integration and the details are enriched through detail enhancement step. In this section, we conduct internal analysis to evaluate the performance gain through integrating each module. Our evaluation is conducted on the MultiPIE dataset where we follow the training and evaluation strategies illustrated above. We start to train from scratch using several convolutional layers without a long-range skip connection and empirically find that it does not converge in practice. Inspired by the VDSR \cite{kim-cvpr16-accurate} method where the output is the combination of the input image and the last layer output, we design a similar structure which is the baseline (denoted as B) of our method. It contains all the 53 convolutional layers and nonlinear activations with a long-range skip connection while the local skip connections are removed. In addition to the baseline configuration, we integrate local residual blocks (i.e., short-range skip connections) to see the performance gain. It is denoted as BL in this study. Note that in this two configurations, we only take the LR blurry face images as input to train the network. In order to evaluate the effectiveness of facial masks, we add them as the input together with the input image. This configuration is denoted as BL+M. Then, we follow the same configuration as BL+M and integrate the dilation into the network, which is denoted as BLD+M. Finally, we add our detail enhancement module and denote it as BLD+M+DE. These configurations indicate how facial masks, local residual blocks, dilation and detail enhancement affect the image quality of the output results. Moreover, we evaluate the effectiveness of each module when the input is an LR image with and without random motion blur, independently. It corresponds to how our method handles both hallucination and deblurring tasks.

\renewcommand{\tabcolsep}{1pt}
\begin{figure}[t]
\begin{center}
\begin{tabular}{ccc}
\includegraphics[width=0.32\linewidth]{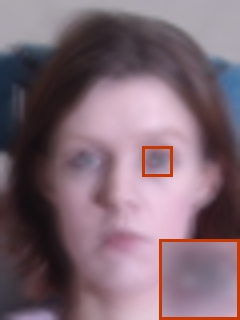}&
\includegraphics[width=0.32\linewidth]{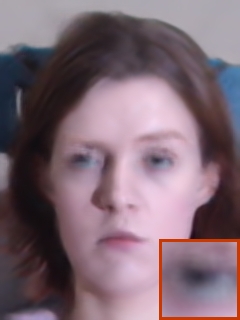}&
\includegraphics[width=0.32\linewidth]{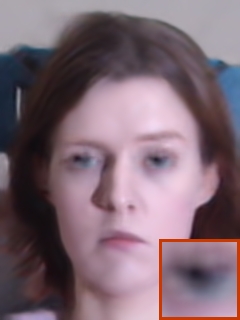}\\
(a) Input (Bic)&(b) B&(c) BL\\
\includegraphics[width=0.32\linewidth]{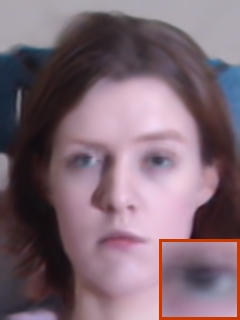}&
\includegraphics[width=0.32\linewidth]{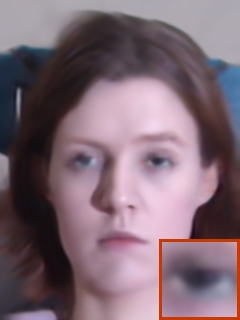}&
\includegraphics[width=0.32\linewidth]{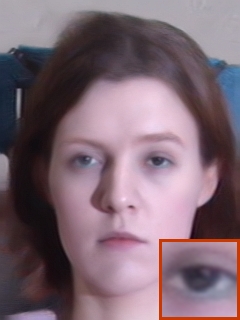}\\
(d) BL+M&(e) BLD+M&(f) BLD+M+DE\\
\end{tabular}
\end{center}
\caption{Visualization of the ablation studies. Figure (a) shows the bicubic upscaled input LR blurry image. Figure (b) shows the baseline network performance and (c) shows the local residual blocks integration on (b). In (d), we retrain (c) using facial masks and generate the result. Meanwhile, we involve dilation in (d) and generate the result shown in (e). The detail enhancement on (e) is shown in (f).}
\label{fig:expA}
\end{figure}

Table \ref{tab:ablation} shows the quantitative evaluation performance of five configurations under PSNR and SSIM metrics. For each configuration, the output images are generated based on the LR input image with and without motion blur, respectively. Then we compute the numerical results and average them to obtain the listed numbers. The quantitative results show that local residual blocks improve the baseline performance and facial masks improve more when adopted as the input for both SR and Deblur scenarios. Meanwhile, the dilation on our FSGN module is effective to predict the output and detail enhancement makes a further improvement. The performance gain is consistent with both PSNR and SSIM metrics. It indicates that facial masks, local residual blocks, dilation and detail enhancement will contribute to the quality of the face images for hallucination and deblurring. We note that the facial masks further improve the performance on the deblurring tasks compared with the hallucination task. It is because the facial masks enable CNN to attend to facial components containing unique structures, which usually diminish on the blurry inputs.

Figure \ref{fig:expA} shows a visual example of these configurations. The input LR blurry face image is shown in (a) and the result generated by the baseline is in (b). The blurry effect still exists around the facial component and little improvement is achieved through local residual blocks integration shown in (c). However, when using facial masks, we notice that the blur around facial component is effectively reduced shown in (d). Furthermore, the dilation integration makes a further improvement (i.e, the right eye region in the close-up) shown in (e). The result generated by all the modules is shown in (f) where local details are transferred from HR exemplar images to (e) in the detail enhancement step.

\def\pp{\hspace{0mm}}
\renewcommand{\tabcolsep}{2pt}
\begin{table}[t]
\caption{The evaluation of the Multi-PIE dataset with the state-of-the-art methods.}
\centering
       \begin{tabular}{cccc}
        \toprule
        &PSNR&SSIM&Similarity\\
        &SR / Deblur&SR / Deblur&SR / Deblur\\
        \midrule
        Bicubic&32.43 / 23.58&0.89 / 0.81&0.92 / 0.87\\
        FHTP \cite{liu-ijcv07-FH}&30.13 / 23.13&0.82 / 0.77&0.90 / 0.86\\
        SFH \cite{yang-cvpr13-sfh}&31.60 / 23.65&0.86 / 0.79&0.91 / 0.86\\
        SRCNN \cite{dong-pami2016-image}&33.89 / 23.73&0.90 / 0.82&0.94 / 0.90\\
        SRCSC \cite{shuhang-iccv15-csc}&33.95 / 23.82&0.90 / 0.82&0.95 / 0.91\\
        SRResNet \cite{ledig-cvpr17-photo}&34.10 / 23.95&0.90 / 0.81&0.96 / 0.92\\
        VDSR \cite{kim-cvpr16-accurate}&34.62 / 24.33&0.91 / 0.81&0.97 / 0.92\\
        DFE \cite{pan-eccv14-dfe}&31.53 / 25.26&0.87 / 0.85&0.94 / 0.94\\
        RBF \cite{xu-iccv17-learning}&30.05 / 24.73&0.86 / 0.77&0.93 / 0.94\\
        Ours&\textbf{34.93} / \textbf{25.75}&\textbf{0.92} / \textbf{0.86}&\textbf{0.98} / \textbf{0.96}\\
        \bottomrule
       \end{tabular}
\label{tab:exp1}
\vspace{3mm}
\caption{The evaluation of the PubFig dataset with the state-of-the-art methods.}
\centering
       \begin{tabular}{cccc}
        \toprule
        &PSNR&SSIM&Similarity\\
        &SR / Deblur&SR / Deblur&SR / Deblur\\
        \midrule
        Bicubic&29.55 / 22.79&0.86 / 0.83&0.89 / 0.84\\
        FHTP \cite{liu-ijcv07-FH}&26.56 / 22.51&0.71 / 0.79&0.87 / 0.83\\
        SFH \cite{yang-cvpr13-sfh}&28.51 / 22.70&0.82 / 0.81&0.88 / 0.84\\
        SRCNN \cite{dong-pami2016-image}&31.03 / 22.85&0.88 / 0.84&0.91 / 0.86\\
        SRCSC \cite{shuhang-iccv15-csc}&31.15 / 23.13&0.88 / 0.85&0.92 / 0.87\\
        SRResNet \cite{ledig-cvpr17-photo}&31.23 / 23.21&0.88 / 0.85&0.94 / 0.87\\
        VDSR \cite{kim-cvpr16-accurate}&31.67 / 23.46&0.89 / 0.86&0.94 / 0.86\\
        DFE \cite{pan-eccv14-dfe}&28.74 / 23.95&0.81 / 0.87&0.89 / 0.88\\
        RBF \cite{xu-iccv17-learning}&28.43 / 23.31&0.80 / 0.86&0.88 / 0.87\\
        Ours&\textbf{31.86} / \textbf{24.12}&\textbf{0.90} / \textbf{0.89}&\textbf{0.96} / \textbf{0.90}\\
        \bottomrule
       \end{tabular}
\label{tab:exp2}
\end{table}

\renewcommand{\tabcolsep}{1pt}
\begin{figure*}[t]
\begin{center}
\begin{tabular}{cccc}
\includegraphics[width=0.24\linewidth]{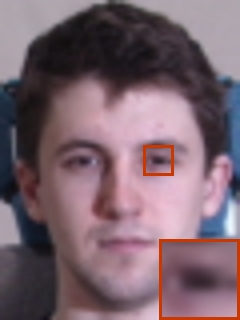}&
\includegraphics[width=0.24\linewidth]{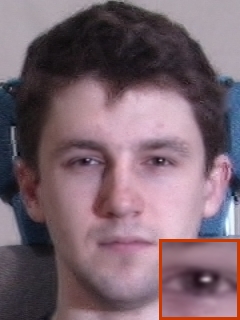}&
\includegraphics[width=0.24\linewidth]{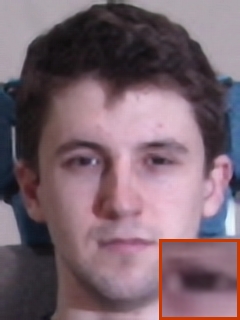}&
\includegraphics[width=0.24\linewidth]{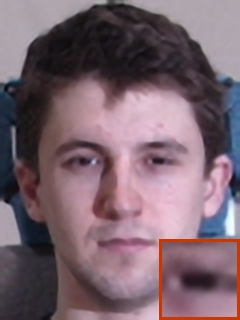}\\
(a) Input (Bic)&(b) SFH \cite{yang-cvpr13-sfh}&(c) SRCNN \cite{dong-pami2016-image}&(d) SRCSC \cite{shuhang-iccv15-csc}\\
32.78 / 0.88& 32.81 / 0.86& 34.65 / 0.90& 35.59 / 0.91\\
\includegraphics[width=0.24\linewidth]{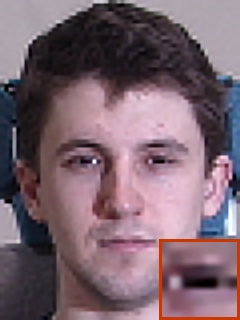}&
\includegraphics[width=0.24\linewidth]{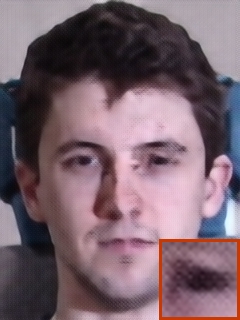}&
\includegraphics[width=0.24\linewidth]{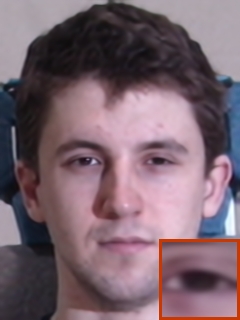}&
\includegraphics[width=0.24\linewidth]{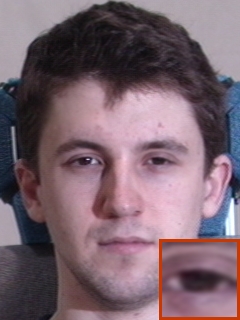}\\
(e) DFE \cite{pan-eccv14-dfe}&(f) RBF \cite{xu-iccv17-learning}&(g) Ours&(h) Ground Truth\\
30.12 / 0.85& 29.84 / 0.82& \textbf{36.77} / \textbf{0.93} & PSNR / SSIM\\
\end{tabular}
\end{center}
\caption{Qualitative evaluations on the Multiple Dataset. (a) shows the bicubic upsampled input LR image without motion blur. (b)-(f) show comparison of the results. (g) denotes our result. (h) denotes the ground truth image.}
\label{fig:exp1}
\end{figure*}

\begin{figure*}[t]
\begin{center}
\begin{tabular}{cccc}
\includegraphics[width=0.24\linewidth]{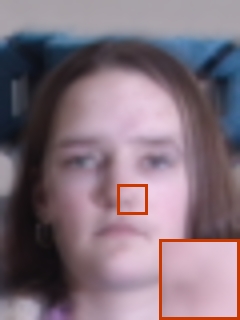}&
\includegraphics[width=0.24\linewidth]{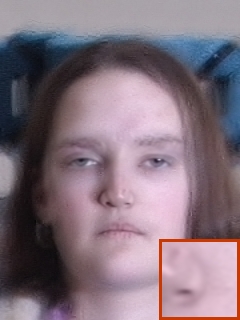}&
\includegraphics[width=0.24\linewidth]{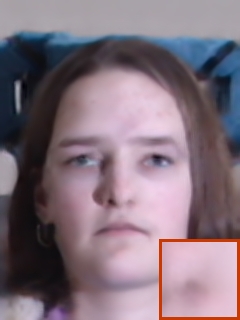}&
\includegraphics[width=0.24\linewidth]{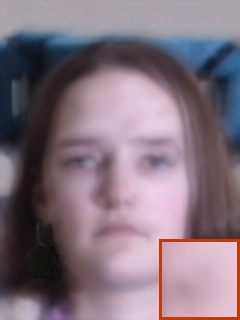}\\
(a) Input (Bic)&(b) SFH \cite{yang-cvpr13-sfh}&(c) VDSR \cite{kim-cvpr16-accurate}&(d) SRResNet \cite{ledig-cvpr17-photo}\\
23.60 / 0.81& 23.51 / 0.78& 24.98 / 0.84& 23.68 / 0.81\\
\includegraphics[width=0.24\linewidth]{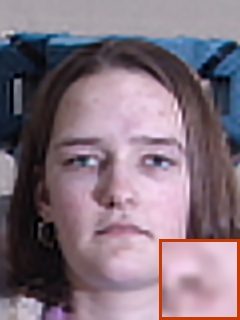}&
\includegraphics[width=0.24\linewidth]{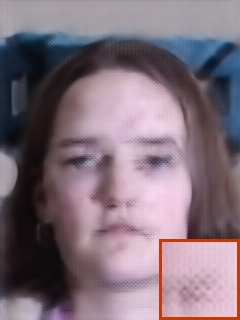}&
\includegraphics[width=0.24\linewidth]{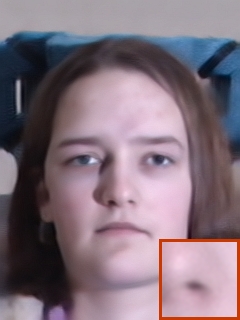}&
\includegraphics[width=0.24\linewidth]{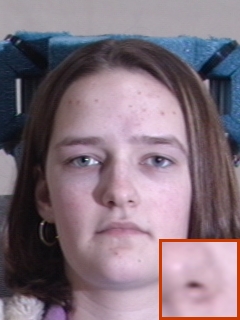}\\
(e) DFE \cite{pan-eccv14-dfe}&(f) RBF \cite{xu-iccv17-learning}&(g) Ours&(h) Ground Truth\\
25.52 / 0.84 & 24.33 / 0.75 & \textbf{25.81} / \textbf{0.86} & PSNR / SSIM\\
\end{tabular}
\end{center}
\caption{Qualitative evaluation on the Multiple Dataset. (a) denotes the LR input blurry image with bicubic upsampling.}
\label{fig:exp2}
\end{figure*}

\begin{figure*}[t]
\begin{center}
\begin{tabular}{cccc}
\includegraphics[width=0.24\linewidth]{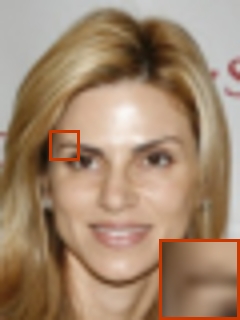}&
\includegraphics[width=0.24\linewidth]{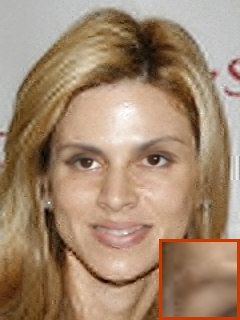}&
\includegraphics[width=0.24\linewidth]{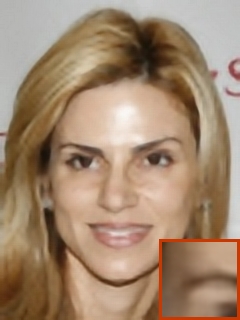}&
\includegraphics[width=0.24\linewidth]{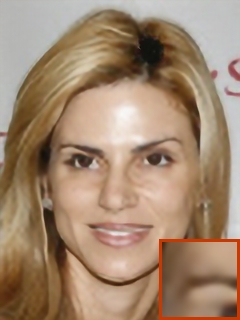}\\
(a) Input (Bic)&(b) SFH \cite{yang-cvpr13-sfh}&(c) SRCNN \cite{dong-pami2016-image}&(d) SRResNet \cite{ledig-cvpr17-photo}\\
31.20 / 0.84& 30.39 / 0.81& 33.37 / 0.88& 32.99 / 0.88\\
\includegraphics[width=0.24\linewidth]{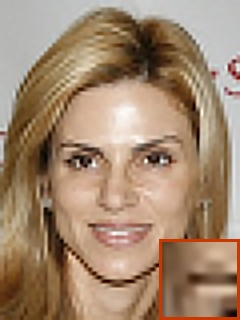}&
\includegraphics[width=0.24\linewidth]{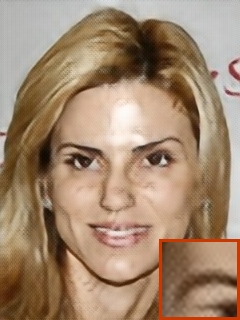}&
\includegraphics[width=0.24\linewidth]{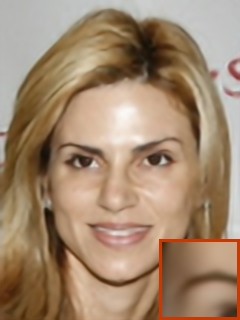}&
\includegraphics[width=0.24\linewidth]{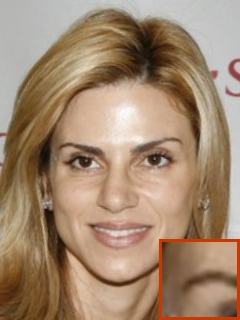}\\
(e) DFE \cite{pan-eccv14-dfe}&(f) RBF \cite{xu-iccv17-learning}&(g) Ours&(h) Ground Truth\\
28.85 / 0.81& 29.05 / 0.78& \textbf{34.06} / \textbf{0.90} & PSNR / SSIM\\
\end{tabular}
\end{center}
\caption{Qualitative evaluation on the PubFig Dataset. (a) denotes the LR input image without motion blur which is generated by bicubic upsampling.}
\label{fig:exp3}
\end{figure*}

\begin{figure*}[t]
\begin{center}
\begin{tabular}{cccc}
\includegraphics[width=0.24\linewidth]{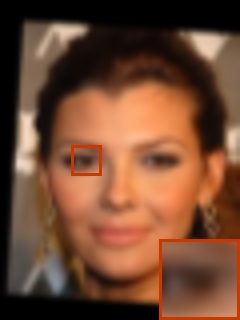}&
\includegraphics[width=0.24\linewidth]{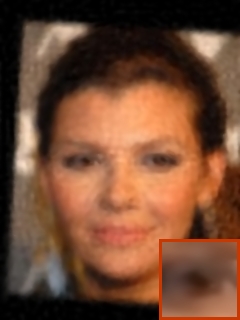}&
\includegraphics[width=0.24\linewidth]{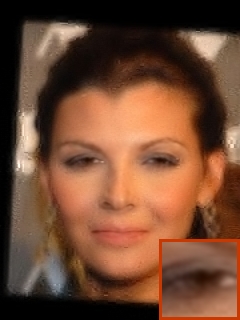}&
\includegraphics[width=0.24\linewidth]{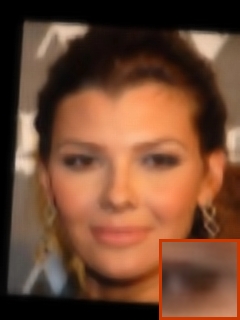}\\
(a) Input (Bic)&(b) FHTP \cite{liu-ijcv07-FH}&(c) SFH \cite{yang-cvpr13-sfh}&(d) SRCNN \cite{dong-pami2016-image}\\
22.53 / 0.85& 22.22 / 0.79& 22.60 / 0.81& 22.62 / 0.86\\
\includegraphics[width=0.24\linewidth]{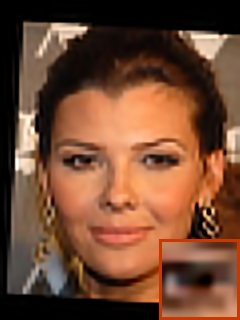}&
\includegraphics[width=0.24\linewidth]{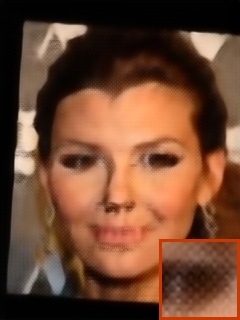}&
\includegraphics[width=0.24\linewidth]{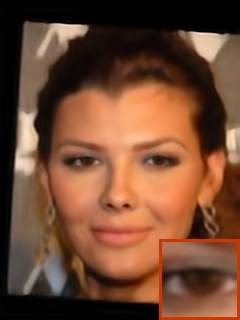}&
\includegraphics[width=0.24\linewidth]{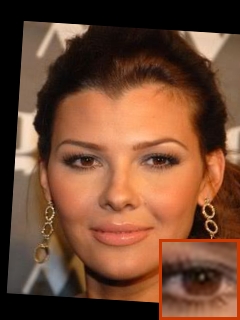}\\
(e) DFE \cite{pan-eccv14-dfe}&(f) RBF \cite{xu-iccv17-learning}&(g) Ours&(h) Ground Truth\\
23.04 / 0.88 & 21.89 / 0.76 & \textbf{23.32} / \textbf{0.89} & PSNR / SSIM\\
\end{tabular}
\end{center}
\caption{Qualitative evaluation on the PubFig Dataset. (a) denotes the LR input blurry image which is generated by bicubic upsampling.}
\label{fig:exp4}
\end{figure*}

\subsection{Comparisons with the State-of-the-art Methods}
We compare our method with the state-of-the-art methods both quantitatively and qualitatively. Table \ref{tab:exp1} reports the quantitative performance on the Multi-PIE dataset. In addition to the PSNR and SSIM metrics, we also involve the identity similarity \cite{delac-ijcst05i-independent} to measure how the results generated by different methods resemble the ground truth images. We use the ground truth training images to construct a PCA projection matrix which projects the results and corresponding HR images. After projection, we compute the cosine distance between each result and the corresponding ground truth image. This identity similarity is set to quantitatively measure image quality from the perspective of face recognition.

Table \ref{tab:exp1} shows that the bicubic interpolation achieves higher PSNR values than existing FH methods (i.e., FHTP and SFH) under both SR and deblur tasks. This is because FH methods establish HR correspondences through image alignment which is based on empirical features such as SIFT \cite{liu-pami11-siftflow}. As the resolution of the input image is low, empirical features cannot accurately locate HR correspondences. It leads to the mismatch and incorrect facial details will be transferred. As a result, around facial component areas, we will find the distortion of the shape, shifting of the location or the change of the lightness, as shown in Figure \ref{fig:exp1} (b), Figure \ref{fig:exp2} (b) and Figure \ref{fig:exp4}(b). These artifacts deteriorate the image quality. In the FH task, the SRCNN, SRCSC and SRResNet methods achieve high PSNR values due to their global optimization scheme. However, blur occurs around high-frequency facial structures including eyes, noses, and mouth, which limits the image quality as well. Meanwhile, their performance decreases on the deblurring task. In comparison, the DFE and RBF methods are effective to handle motion blur while limiting their performance in hallucination. Different from existing methods which handle FH and FD independently, our method consists of a unified framework to jointly hallucinate and deblur face images. It recovers the original image content in both low and high frequencies, which enables the similarity of global appearance and local details between the output image and the ground truth. The evaluation of the PubFig dataset shows the similar performance in Table \ref{tab:exp2}. It indicates our method is effective to overcome real-world input variations. Table \ref{tab:exp1} and Table \ref{tab:exp2} show that our method performs favorably against the state-of-the-art FH, image SR and FD methods.

Besides quantitative evaluation, we also evaluate our method visually on the benchmarks. We show the qualitative comparison from Figure \ref{fig:exp1} to Figure \ref{fig:exp4}.
In Figure \ref{fig:exp1}, we evaluate the proposed algorithm on the Multi-PIE dataset using the input LR image without motion blur.
The result generated by SFH is shown in (b) contains light dots on the right eye, which is different from the ground truth.
This is because SFH selects the most similar component from the dataset and transfer its gradient to recover high-frequency details.
However, the facial component correspondence cannot be well established in LR. In this case, gradient transfer leads to the dissimilar generation of the facial structure.
Another visual result is shown in Figure \ref{fig:exp3} (b) where the lighting, shape and position of the facial components are different from those in Figure \ref{fig:exp3} (h) although they look similar.
It also indicates that erroneous gradient transfer brings artifacts due to inaccurate correspondence establishment.
In addition, noise is included due to incorrect matching around the mouth region. As the PubFig dataset is taken in the real world condition and the training dataset is taken in the lab controlled environment.
The component matching is not as accurate as that in Multi-PIE. It brings more artifacts on the generated results.

The results by the representative CNN-based methods are shown in Figure \ref{fig:exp1} (c) and Figure \ref{fig:exp1} (d).
Although the recovered images have high PSNR and SSIM values compared to that by SFT method, these methods are not effective to capture high-frequency facial details.
The structures around facial components (i.e., eyes, nose and mouth) are blurry and details are missing.
In addition, the results generated by VDSR and SRResNet shown in Figure \ref{fig:exp2} (c) and Figure \ref{fig:exp2} (d) show the similar performance.
This indicates that CNN methods for general image SR are less effective to preserve details on face images.
To solve this problem, we generate the base image through CNN prediction and enhance details via HR exemplar images.
The base image contains the low-frequency facial structures similar to the existing CNN based methods.
Meanwhile, we synthesize fine-grained structures from HR exemplar images and transfer their high-frequency details back to the base image for enhancement.
Our two-stage scheme enables our results are similar to the ground truth in both global appearance and local details shown in Figure \ref{fig:exp1} (g) and Figure \ref{fig:exp3} (g).
The proposed algorithm achieves favorable performance under numerical evaluations as well as visual perception.

\begin{figure}[t]
\begin{center}
\begin{tabular}{ccc}
\includegraphics[width=0.32\linewidth]{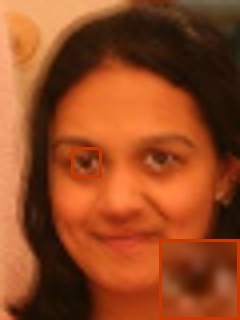}&
\includegraphics[width=0.32\linewidth]{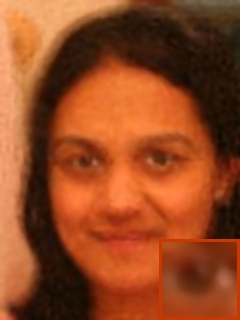}&
\includegraphics[width=0.32\linewidth]{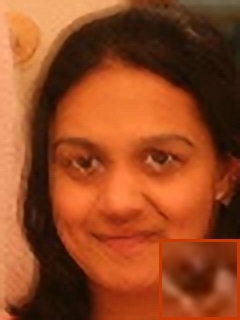}\\
(a) Input &(b) FHTP \cite{liu-ijcv07-FH}&(c) SRCNN \cite{dong-pami2016-image}\\
\includegraphics[width=0.32\linewidth]{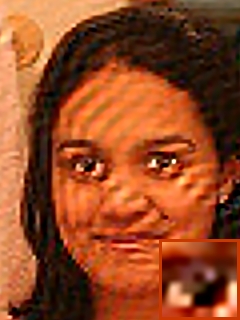}&
\includegraphics[width=0.32\linewidth]{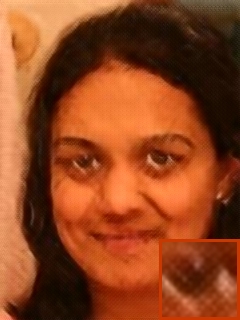}&
\includegraphics[width=0.32\linewidth]{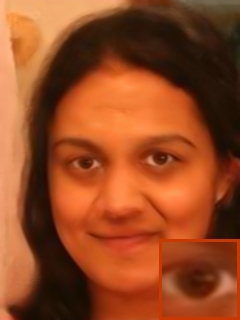}\\
(d) DFE \cite{pan-eccv14-dfe}&(e) RBF \cite{xu-iccv17-learning}&(f) Ours\\
\end{tabular}
\end{center}
\caption{Qualitative evaluation on a real blurry face image. (a) denotes the LR input blurry image generated by bicubic upsampling.}
\label{fig:exp5}
\end{figure}

Figure \ref{fig:exp2} and Figure \ref{fig:exp4} show the qualitative evaluation on the blurry Multi-PIE and PubFig datasets, respectively. The input is an LR face image with random motion blur. It limits the performance of existing FH and image SR results shown in Figure \ref{fig:exp2} (b)-(d) and Figure \ref{fig:exp4} (b)-(d). The exemplar-based face deblurring method DFE selects a suitable exemplar and transfers the gradient into the blurry input image. It is effective to retain the low-frequency structure while limits its performance to restore the facial details. As shown in Figure \ref{fig:exp2} (e) and Figure \ref{fig:exp4} (e), the artifacts appear on the whole image. Meanwhile, the results generated by the GAN network cannot reduce the artifacts and deteriorates the facial structure as shown in Figure \ref{fig:exp2} (f) and Figure \ref{fig:exp4} (f). They aim to solve extremely low-resolution face images (e.g., 20$\times$20) with ambiguous facial components, which are similar to noise. The GAN loss function will introduce fake details thus degrading the quality of the restored face images. The identity of their output is usually not preserved compared with the ground truth. In comparison, our method first generates the base image to reduce the blurry effect and further enhance the structure details. It can jointly handle the hallucination and deblurring tasks where it performs favorably against existing FH and FD methods quantitatively and qualitatively.

Besides evaluation of the standard benchmarks using synthetic motion blur kernels, we also evaluate on the blurry face images in the real world condition. Figure \ref{fig:exp5} (a) shows an example where the input is a real blurry face image from \cite{Lai-CVPR-2016}. It fails existing exemplar-based FH and FD methods to establish an accurate correspondence between the input and the exemplar, which brings artifacts shown in (b) and (d). Meanwhile, the CNN based methods are not effective to reduce the blur shown in (c) and (e). Different from existing methods, our method first generates a base image to facilitate exemplar matching and then performs detail enhancement on the base image. It accurately transfers details from the exemplar to the base image shown in (f), which indicates that our method is effective to reduce real blurry face images.

\subsection{Computational Cost}
We evaluate the time cost of each method to generate an output image. All the evaluations are conducted on a
PC with an i7 3.6GHz CPU and a Tesla K40c GPU. Table \ref{tab:time} shows the time cost of each method. We observe that the exemplar-based methods (i.e., FHTP, SFH, DEF) consume much time cost, which is mainly because of the querying on the exemplar dataset. In comparison, the CNN based methods (i.e., SRCNN, SRResNet, VDSR, RBF) take less time for an end-to-end prediction. Our method consists of the CNN prediction and exemplar-based searching, which takes more time than end-to-end CNN prediction while still performs favorably against exemplar-based methods.

\def\pp{\hspace{0mm}}
\renewcommand{\tabcolsep}{1pt}
\begin{table}[t]
\caption{The time cost of comparing methods to generate an output $320\times240$ image on the benchmarks.}
\centering
       \begin{tabular}{cccc}
        \toprule
        Methods&Time (sec)&Methods&Time (sec)\\
        \midrule
        FHTP&98.9&SFH&245.1\\
        SRCNN&4.36&SRCSC&43.56\\
        SRResNet&6.5&VDSR&5.7\\
        DFE&162.4&RBF&4.2\\
        Ours&95.3&&\\
        \bottomrule
       \end{tabular}
\label{tab:time}
\end{table}
\vspace{-4mm}

\subsection{Limitation}
\begin{figure}[t]
\begin{center}
\begin{tabular}{ccc}
\includegraphics[width=0.32\linewidth]{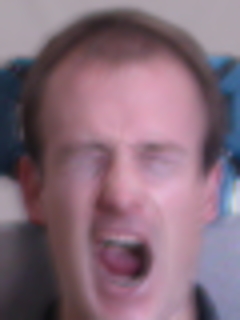}&
\includegraphics[width=0.32\linewidth]{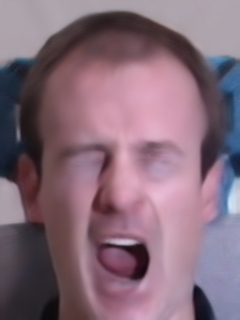}&
\includegraphics[width=0.32\linewidth]{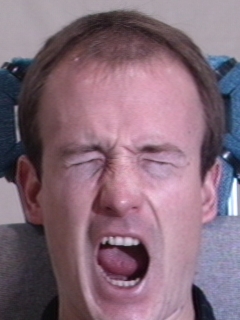}\\
(a) Input&(b) Ours&(c) Ground Truth\\
\end{tabular}
\end{center}
\caption{Limitations of the proposed method. (a) denotes the LR input image which is generated by bicubic upsampling.}
\label{fig:limit}
\end{figure}

The proposed algorithm is less effective when the structures around the facial component are not available or significantly different from the training images.
In such cases, the proposed algorithm would reduce to a conventional CNN-based image restoration algorithm as the facial components do not help the estimation. Figure \ref{fig:limit} shows an example where our method is not able to recover clear face images.

\renewcommand{\tabcolsep}{1pt}

\section{Concluding Remarks}
We propose an effective algorithm to jointly hallucinate and deblur face images.
With the guidance of facial components, we develop an FSGN to remove blur and restore the major structures of face images.
To recover realistic faces, we develop a detail enhancement algorithm by high-resolution exemplars. Our analysis shows that the proposed method is able to generate high-resolution faces from blurry LR face images.
Extensive experimental results demonstrate that our method performs favorably against the state-of-the-art approaches.

\section*{Acknowledgments}
\vspace{-2mm}
This work has been supported in part by the NSF CAREER (No. 1149783), NSF of China (No. 61872421), and NSF of Jiangsu Province (No. BK20180471).

\bibliographystyle{spmpsci}
\bibliography{ref}
\end{document}